\documentclass[10pt,twocolumn,letterpaper]{article}

\usepackage[pagenumbers]{cvpr} 

\usepackage{graphicx}
\usepackage{caption}
\usepackage{amsmath, amsfonts, amssymb}
\usepackage[table,xcdraw]{xcolor}
\usepackage[normalem]{ulem}
\useunder{\uline}{\ul}{}
\usepackage{pifont}
\usepackage{float}
\usepackage[accsupp]{axessibility}  

\definecolor{cvprblue}{rgb}{0.21,0.49,0.74}
\usepackage[pagebackref,breaklinks,colorlinks,allcolors=cvprblue]{hyperref}

\def\paperID{****} 
\def\confName{CVPR}
\def\confYear{2026}

\title{\textsc{Mobile-VTON}: High-Fidelity On-Device Virtual Try-On}

\author{ Zhenchen Wan$^{1,2*}$ \quad Ce Chen$^{3*}$ \quad Runqi Lin$^{1}$ \quad Jiaxin Huang$^{2}$ \quad Tianxi Chen$^{3}$ \\ Yanwu Xu$^{4}$ \quad Tongliang Liu$^{1,2\dagger}$ \quad Mingming Gong$^{2,3\dagger}$ \\ [0.6em]
$^{1}$University of Sydney \quad $^{2}$MBZUAI \quad $^{3}$University of Melbourne \quad $^{4}$Google \\  \tt\small \{zwan0681, tongliang.liu\}@sydney.edu.au, \\ \tt\small ce.chen@student.unimelb.edu.au, mingming.gong@unimelb.edu.au \\  \tt\normalsize * Equal contribution \qquad $\dagger$ Corresponding authors}

\begin{document}

\twocolumn[{%
\renewcommand\twocolumn[1][]{#1}%
\maketitle
\begin{center}
    \vspace{-1.0em}
    \centering
    \includegraphics[width=1\textwidth]{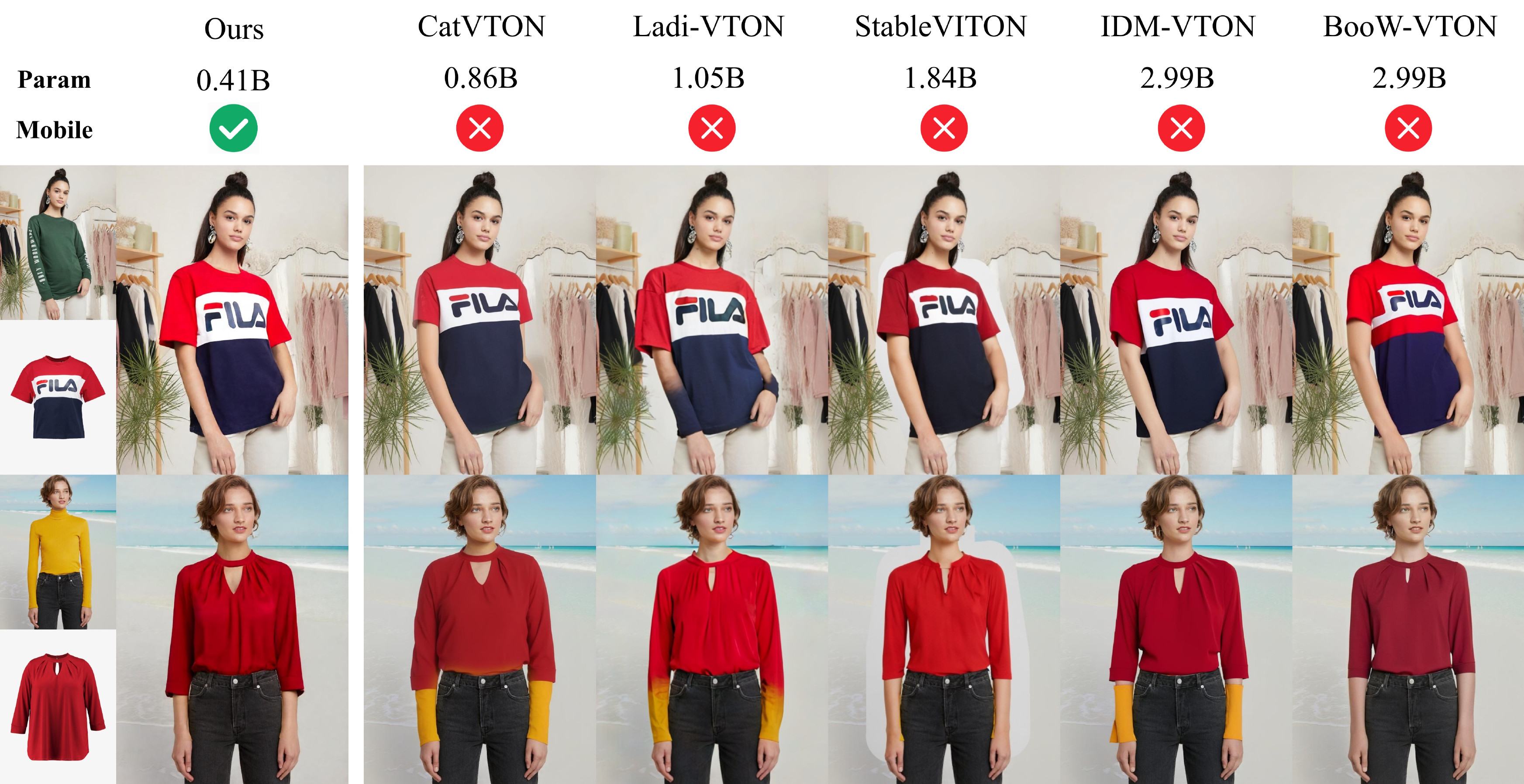}
    \captionsetup{type=figure, skip=-1pt}
    \captionof{figure}{Comparison of virtual try-on methods in terms of model size, mobile compatibility, and visual quality. Our model, with only 415M parameters, achieves competitive visual results while running entirely on mobile devices. The leftmost column shows the input person and garment images. All outputs are generated or super-resolved at $1024 \times 768$ resolution. Zoom in for details.
    }
    \label{fig:cover_image}
\end{center}%
}]

\begin{abstract}

Virtual try-on (VTON) has recently achieved impressive visual fidelity, but most existing systems require uploading personal photos to cloud-based GPUs, raising privacy concerns and limiting on-device deployment. To address this, we present \textsc{Mobile-VTON}, a high-quality, privacy-preserving framework that enables fully offline virtual try-on on commodity mobile devices using only a single user image and a garment image. \textsc{Mobile-VTON} introduces a modular TeacherNet--GarmentNet--TryonNet (TGT) architecture that integrates knowledge distillation, garment-conditioned generation, and garment alignment into a unified pipeline optimized for on-device efficiency. Within this framework, we propose a Feature-Guided Adversarial (FGA) Distillation strategy that combines teacher supervision with adversarial learning to better match real-world image distributions. GarmentNet is trained with a trajectory-consistency loss to preserve garment semantics across diffusion steps, while TryonNet uses latent concatenation and lightweight cross-modal conditioning to enable robust garment-to-person alignment without large-scale pretraining. By combining these components, \textsc{Mobile-VTON} achieves high-fidelity generation with low computational overhead. Experiments on VITON-HD and DressCode at $1024{\times}768$ show that it matches or outperforms strong server-based baselines while running entirely offline. These results demonstrate that high-quality VTON is not only feasible but also practical on-device, offering a secure solution for real-world applications. Code and project page are available at \url{https://zhenchenwan.github.io/Mobile-VTON/}.

\end{abstract}
    
\section{Introduction}
\label{sec:intro}

Virtual Try-On (VTON) technology has emerged as a transformative solution for the fashion and e-commerce industries, enabling users to visualize how garments would look on their bodies without physical trials~\cite{morelli_dress_2022, wan_ted-viton_2024, he_vton_2025, wan2025mft, fele_c-vton_2022, yang_omnivton_2025}. Recent advances in large-scale diffusion models have significantly improved visual fidelity and garment detail preservation~\cite{bai_single_2022, choi_viton-hd_2021, dong_fw-gan_2019, han_viton_2018, minar_cp-vton_2020, wang_toward_2018, yang_towards_2020, yu_vtnfp_2019}. However, these models typically rely on cloud-based GPUs, requiring users to upload personal images for inference. This approach not only introduces latency and energy overhead but also raises serious privacy concerns, particularly under stringent data protection regulations. In response, our goal is to design a fully on-device VTON system that avoids any transmission of user data, while still achieving high-fidelity synthesis.

To bridge this gap, we identify three key challenges that limit efficient on-device deployment and guide the design of our system. (i) Diffusion-based VTON models typically have large parameter sizes, resulting in high memory consumption and latency that exceed the capabilities of mobile NPUs and GPUs. (ii) Garment representations can shift across diffusion timesteps, causing semantic drift and texture distortion. These inconsistencies reduce visual coherence, impair identity preservation, and complicate control over garment attributes such as structure, material, and pose. (iii) Existing approaches often rely on models pre-trained on large-scale image datasets before being fine-tuned for VTON tasks. Without such pretraining, these models struggle to acquire strong garment synthesis capabilities, particularly in low-data or minimal-input settings. This reliance on extensive upstream training pipelines limits the flexibility of lightweight architectures and poses a major obstacle to building standalone on-device systems that can be trained directly on the VTON task.

To address these challenges, we propose \textsc{Mobile-VTON}, a unified, privacy-preserving framework optimized for on-device deployment, built around a TeacherNet--GarmentNet--TryonNet (TGT) architecture. (i) To enable efficient and realistic synthesis under mobile constraints, we introduce a Feature-Guided Adversarial (FGA) Distillation strategy that transfers fine-grained generative  knowledge from a high-capacity teacher while leveraging adversarial supervision to align the student model’s outputs with real-world image distributions. This allows compact student networks to synthesize realistic try-on results under tight mobile compute and memory constraints. (ii) To address semantic instability across diffusion steps, GarmentNet is trained with a trajectory-consistency objective that enforces garment feature invariance over timesteps. This improves robustness to pose and viewpoint changes and reduces semantic drift. (iii) To reduce reliance on large-scale pretraining, we design TryonNet as a garment-aware synthesis module that learns garment–body alignment directly within the virtual try-on task. It incorporates a cross-modal fusion mechanism that jointly encodes person and garment features in latent space, enabling effective training even in data-limited settings.

We evaluate \textsc{Mobile-VTON} on three benchmark datasets: VITON-HD~\cite{choi_viton-hd_2021}, DressCode~\cite{morelli_dress_2022}, and VITON-HD In-the-Wild, which was introduced by BooW-VITON~\cite{zhang_boow-vton_2024} which includes more realistic usage scenarios. All experiments are conducted at a resolution of $1024{\times}768$, using only a single person image and a single garment image as input.  Despite operating in this minimal-input setting and running fully offline on commodity mobile devices, \textsc{Mobile-VTON} achieves performance that is competitive with or superior to state-of-the-art server-based baselines, as measured by perceptual and structural metrics including LPIPS, SSIM, and CLIP-I. These results highlight the practical viability of our approach, demonstrating that high-quality virtual try-on can be achieved entirely on-device, with improved responsiveness and enhanced user privacy. In summary, our contributions are as follows:

\begin{itemize}
    \item We present \textsc{Mobile-VTON}, the first diffusion-based VTON system, to the best of our knowledge, capable of running entirely on commodity mobile devices without requiring any additional user-provided information beyond the garment and person images.
    
    \item We propose a novel TGT framework designed specifically for mobile deployment. We incorporate the proposed FGA Distillation for efficient and realistic generation, a trajectory-consistent GarmentNet to maintain semantic stability across diffusion steps, and a garment-aware TryonNet with perceptual priors, a lightweight adapter, and cross-modal fusion for accurate and texture-preserving garment-to-body alignment. The system is trained from scratch on task-specific data without external pretraining.
    
    \item We demonstrate that \textsc{Mobile-VTON} achieves high visual quality under mobile constraints, matching or outperforming server-based baselines, while ensuring full offline operation and strong privacy guarantees.
\end{itemize}
\section{Related Works}
\label{sec:related}

\vspace{0.5em}
\noindent\textbf{GAN-based VTON.} Classical GAN pipelines~\cite{goodfellow_generative_2014} perform (i) body structure estimation, (ii) garment warping, and (iii) image fusion. VITON-GAN and FW-GAN are early two-stage designs coupling geometric alignment with adversarial refinement~\cite{honda_viton-gan_2019, dong_fw-gan_2019}. Later work improved locality and resolution: InsetGAN’s patch edits~\cite{fruhstuck_insetgan_2022}, HR-VITON’s high-resolution scaling~\cite{lee_high-resolution_2022}, and GP-VTON’s layered parsing/warping for detail preservation~\cite{xie_gp-vton_2023}. ADGAN and CA-GAN target pose and style controllability and attribute-level edits~\cite{men_controllable_2020, kips_ca-gan_2020}, while GarmentGAN and FitGAN broaden category coverage and fit~\cite{raffiee_garmentgan_2020, pecenakova_fitgan_2022}. 

\begin{figure*}[t]
\centering
\includegraphics[width=1\textwidth]{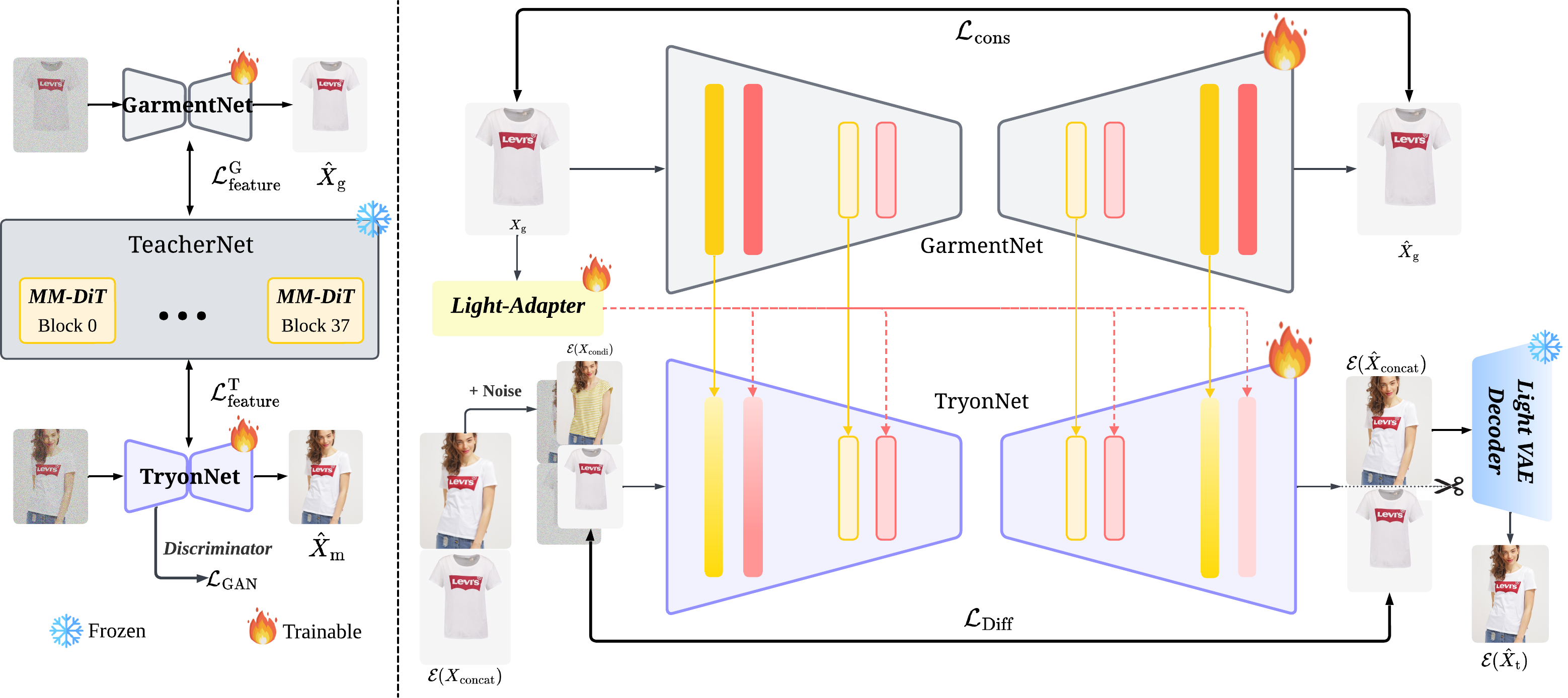}
\caption{\textbf{Training overview of \textsc{Mobile-VTON}.} The left side illustrates our Feature-Guided Adversarial (FGA) Distillation process, where a high-capacity TeacherNet supervises two lightweight student networks: GarmentNet and TryonNet. GarmentNet learns to reconstruct the garment using feature-level distillation loss $\mathcal{L}_{\text{feature}}^{\text{G}}$, while TryonNet is guided by both feature-level loss $\mathcal{L}_{\text{feature}}^{\text{T}}$ and adversarial loss $\mathcal{L}_{\text{GAN}}$. The right side shows the main training pipeline of \textsc{Mobile-VTON}, where TryonNet and GarmentNet are jointly trained with garment-aware supervision. The Light-Adapter injects garment semantics, and the model is optimized using latent concatenation, temporal consistency loss $\mathcal{L}_{\text{cons}}$, and reconstruction loss $\mathcal{L}_{\text{Diff}}$ to generate high-fidelity try-on images.}
\vspace{-1.5em}
\label{fig:Mobile-VTON}
\end{figure*}

\vspace{0.5em}
\noindent\textbf{Mask-based Diffusion-based VTON.} Leveraging diffusion’s fidelity~\cite{ho_denoising_2020}, recent methods frame try-on as exemplar inpainting or latent diffusion with garment conditioning. DCI-VTON uses a warped-garment prior for detail retention~\cite{gou_taming_2023}; StableVITON learns clothing–body correspondences in latent space to reduce explicit warping~\cite{kim_stableviton_2023}. LaDI-VTON standardizes latent pipelines and pseudo-labeling, and OOTDiffusion injects garment features via self-attention~\cite{morelli_ladi-vton_2023, zhu_tryondiffusion_2023}. IDM-VTON adds garment semantics in cross/self-attention for in-the-wild fidelity, while TED-VITON adapts DiT with a garment adapter and text-preservation constraints for crisp logos and prints~\cite{choi_improving_2024, wan_ted-viton_2024}. StreetTryOn contributes an in-the-wild benchmark and a diffusion+DensePose framework for unpaired training~\cite{cui_street_2024}.

\vspace{0.5em}
\noindent\textbf{Mask-free Diffusion-based VTON.} To avoid errors from clothing masks, mask-free diffusion localizes try-on regions implicitly. BooW-VTON learns from large-scale unpaired data without explicit inpainting masks~\cite{zhang_boow-vton_2024}. MF-VITON achieves high fidelity from only a person image and a garment~\cite{wan_mf-viton_2025}. Any2AnyTryon expands controllability and category coverage via adaptive position embeddings and a large open-source garment set~\cite{guo_any2anytryon_2025}. Concurrent efforts explore training-free universality (OmniVTON) and direct garment–body correspondence within diffusion, targeting more general and robust try-on~\cite{yang_omnivton_2025, wan_incorporating_2025}.

\section{Methodology}
\label{sec:method}

\subsection{Model Overview}

As illustrated in Fig.~\ref{fig:Mobile-VTON}, our proposed \textsc{Mobile-VTON} framework introduces a novel modular TeacherNet--GarmentNet--TryonNet (TGT) architecture tailored for high-fidelity, on-device virtual try-on. Each component in the TGT design plays a distinct and complementary role. TeacherNet serves as a high-capacity reference model that guides student training. GarmentNet generates garment-conditioned intermediate features while preserving semantic and temporal consistency. TryonNet synthesizes the final try-on image by aligning person and garment representations through deep fusion.

At the core of this architecture is the Feature-Guided Adversarial (FGA) Distillation strategy, which enables lightweight student networks to approximate the performance of the teacher model under mobile hardware constraints. FGA integrates two objectives: (i) a feature-level score matching loss that distills generative knowledge from TeacherNet into student U-Nets, enhancing texture fidelity and generation stability; and (ii) an adversarial supervision term that improves photorealism by aligning the student's output distribution with that of real images.

\vspace{0.5em}
\noindent\textbf{TeacherNet} is built upon Stable Diffusion 3.5 Large~\cite{esser_scaling_2024} and remains frozen during training.
It serves as a high-capacity generative model that provides supervisory signals for guiding lightweight student networks. Instead of performing explicit pixel-level regression, TeacherNet is treated as a score function oracle that provides gradient-based supervision. During training, given an input pair consisting of a person image $\mathbf{X}_p$ and a garment image $\mathbf{X}_g$, we first encode it into a latent space using a shared VAE encoder $\mathcal{E}$. Here, $\mathbf{X}_p$ is used to supervise the student TryonNet, while $\mathbf{X}_g$ provides supervision for the student GarmentNet. Gaussian noise is then added at a random timestep $t$, and the resulting noisy latent $\tilde{\mathbf{z}}^{(t)}$ is passed to the frozen teacher denoiser $D_t$. The TeacherNet outputs denoising predictions and provides score estimates $s_{\text{true}}(\tilde{\mathbf{z}}^{(t)}, t)$, representing the gradients of the log-probability with respect to the input noise. These scores serve as ground-truth supervision for guiding the student networks.

\vspace{0.5em}
\noindent\textbf{Light-UNets} implement the student-side architecture for this distillation framework. They consist of two lightweight networks, GarmentNet and TryonNet, adapted from the SnapGen architecture~\cite{chen_snapgen_2025} for efficient on-device inference. To match the conditioning capabilities of the teacher, we integrate mobile-optimized attention and dual text encoders (CLIP-G~\cite{radford_learning_2021} and CLIP-L~\cite{radford_learning_2021}). Both GarmentNet and TryonNet operate on noisy latent representations $\tilde{\mathbf{z}}^{(t)}$ at various diffusion timesteps $t$, producing denoised estimates and corresponding score functions $s_{\text{fake}}(\tilde{\mathbf{z}}^{(t)}, t)$ that represent the student’s learned distribution. These outputs define the ``fake" distribution in the distillation setup and are directly compared to the teacher’s score estimates using a DMD2-style loss~\cite{yin_improved_2024}. This approach eliminates the need for trajectory-level regression and supports lightweight generation without compromising visual fidelity.

\vspace{0.5em}
\noindent\textbf{Light-Adapter} is the unified image-to-text conditioning module. It adopts the decoupled cross-attention mechanism from IP-Adapter~\cite{ye_ip-adapter_2023}. For improved mobile efficiency, we replace the large CLIP visual encoder with DINOv2-base~\cite{oquab_dinov2_2024}, a compact yet semantically rich alternative. 
Given a garment image $\mathbf{X}_g$, visual features are extracted via DINOv2 and projected into key and value tensors using learnable linear layers:
\begin{equation}
    (\mathbf{K}_d, \mathbf{V}_d) = \text{Proj}(\text{DINOv2}(\mathbf{X}_g)).
\end{equation}
These tensors encode the garment’s visual semantics and are used as conditioning inputs for TryonNet during cross-attention, providing detailed guidance for garment-to-person alignment and synthesis.

\subsection{Feature-Guided Adversarial Distillation (FGA)}
\label{sec:method:distillation}

Inspired by DMD2~\cite{yin_improved_2024}, we propose a supervision strategy that combines two complementary objectives: a feature-level distillation loss and an adversarial realism loss. Together, they enable the student networks to generate high-quality, realistic outputs while remaining lightweight enough for mobile deployment.

\vspace{0.5em}
\noindent\textbf{Feature-Level Distillation.}
To align the student’s generative behavior with that of the teacher, we adopt a score-based distillation objective. Rather than regressing pixel values or denoised outputs directly, we align the score functions predicted by the teacher and student networks at each diffusion timestep. At a given timestep $t$, the input image (either garment or person) is encoded into latent space via a shared VAE encoder $\mathcal{E}$, and Gaussian noise is added to produce the noisy latent $\tilde{\mathbf{z}}^{(t)}$. This latent is then passed through both the teacher denoiser $D_t$ and the student denoiser $D_s$ to obtain their respective score estimates: $s_{\text{true}}(\tilde{\mathbf{z}}^{(t)}, t)$ from TeacherNet and $s_{\text{fake}}(\tilde{\mathbf{z}}^{(t)}, t)$ from Light-UNets. The student is trained by minimizing the $\ell_2$ distance between these two score functions, defined as:
\begin{equation}
\mathcal{L}_{\text{feature}} =
\mathbb{E}_{t}
\left\|
s_{\text{true}}(\tilde{\mathbf{z}}^{(t)}, t)
-
s_{\text{fake}}(\tilde{\mathbf{z}}^{(t)}, t)
\right\|_2^2.
\label{eq:dmg_final}
\end{equation}
This score-based supervision allows the lightweight student networks to approximate the teacher’s distributional behavior in latent space, thereby improving sample realism and generation stability under mobile constraints.

\vspace{0.5em}
\noindent\textbf{Adversarial Realism Enhancement.}To further improve visual fidelity, we introduce an adversarial loss that encourages the student model to generate perceptually natural images. A lightweight discriminator $D$ is trained to distinguish real person images $X$ from generated images $\hat{X}$. TryonNet is optimized to produce outputs that can fool the discriminator, resulting in more photorealistic results. The adversarial loss is defined as:
\begin{equation}
\mathcal{L}_{\text{GAN}} = \mathbb{E}_{X \sim \mathcal{R}}[\log D(X)] + \mathbb{E}_{\hat{X} \sim \mathcal{G}}[\log (1 - D(\hat{X}))],
\label{eq:gan}
\end{equation}
where $\mathcal{R}$ and $\mathcal{G}$ represent the distributions of real and generated images, respectively. This realism-oriented supervision complements the distillation loss by guiding the student model toward sharper and more coherent outputs.

\subsection{Trajectory-Consistent GarmentNet (TCG)}
\label{sec:method:trajectory}

Unlike conventional diffusion-based generators, GarmentNet in our framework is designed not for image synthesis but for extracting consistent garment features across diffusion timesteps. During training, we observe that naively perturbing garment inputs with noise often leads to semantic drift. Specifically, the garment representation may shift across timesteps, resulting in unstable textures, distorted shapes, or inconsistent conditioning signals. These issues reduce reliability and coherence in the final try-on results.

To address this, we introduce a trajectory consistency constraint that enforces temporal stability in the garment encoding process. Rather than injecting noise during training, we apply the diffusion process deterministically at each timestep $t \in [1, T]$ and require the model to reconstruct the original garment image consistently across all steps. This is formalized as the following objective:
\begin{equation}
    \mathcal{L}_{\text{cons}} = 
    \mathbb{E}_{t \sim [1, T]}
    \left[ \left\| \hat{X}_g^{(t)} - X_g \right\|_2^2 \right],
    \label{eq:cons}
\end{equation}
where $\hat{X}_g^{(t)}$ is the predicted garment reconstruction at timestep $t$, and $X_g$ is the original garment input.
By minimizing this loss, the network is encouraged to produce garment features that remain stable along the diffusion trajectory. This temporal regularization improves the consistency of garment semantics, enabling more robust alignment and rendering in the downstream TryonNet module.

\subsection{Garment-Aware TryonNet}
\label{sec:method:tryon}

The TryonNet module is responsible for synthesizing the final virtual try-on image by aligning garment and person representations through deep fusion. Building on ideas from IDM-VTON~\cite{choi_improving_2024} and CatVTON~\cite{chong_catvton_2024}, our design enhances garment-to-body interaction by integrating multi-level garment features and conditioning priors throughout the generation pipeline. Unlike previous methods, where TryonNet-style modules are typically initialized from large-scale pretrained models (e.g., text-to-image diffusion), our TryonNet is trained from scratch on the downstream virtual try-on task. In this pretraining-free setting, the model lacks prior exposure to garment semantics, which may impair conditioning quality. To address this, we follow a strategy inspired by CatVTON~\cite{chong_catvton_2024}, where garment and person images are spatially concatenated into a unified input. This design explicitly injects garment information into the synthesis pathway, enabling the model to learn alignment directly.

\vspace{0.5em}
\noindent\textbf{Latent Concatenation (LC).}  
To promote accurate spatial alignment between the person and the garment, we first construct a height-wise concatenated image $X_{\text{concat}} = \text{Concat}_{\text{height}}(X_p, X_g)$, and obtain its clean latent representation via a shared VAE encoder as $\mathcal{E}(X_{\text{concat}})$, which is then perturbed with diffusion noise to form the noisy latent $\mathbf{z}_{\text{concat}}$. 
In parallel, we introduce an additional reference input $X_{\text{condi}} = \text{Concat}_{\text{height}}(X_t, X_g)$, which is similarly encoded to $\mathcal{E}(X_{\text{condi}})$ to guide the model in preserving both person identity and garment appearance. 
We then concatenate the two latent codes along the batch dimension to form the final input to TryonNet: $\mathbf{z}_{\text{input}} = \text{Concat}_{\text{batch}}(\mathbf{z}_{\text{concat}}, \mathcal{E}(X_{\text{condi}}))$. This combined latent, along with fused conditioning tokens $(\mathbf{K}_{\text{all}}, \mathbf{V}_{\text{all}})$, is processed through all denoising steps to produce a refined latent representation $\mathcal{E}(\hat{X}_{\text{concat}})$.
Before decoding, we crop the top region of $\mathcal{E}(\hat{X}_{\text{concat}})$ to isolate the person-relevant features, denoted as $\mathcal{E}(\hat{X}_t) = \text{Crop}_{\text{top}}(\mathcal{E}(\hat{X}_{\text{concat}}))$, and decode it via a lightweight VAE decoder $\mathcal{D}(\cdot)$~\cite{chen_snapgen_2025} to reconstruct the final output image as $\hat{X}_t = \mathcal{D}(\mathcal{E}(\hat{X}_t))$.

\vspace{0.5em}
\noindent\textbf{Garment-Aware Reconstruction.}  
To preserve garment details and spatial coherence, we define a reconstruction loss on the full concatenated image. The model is supervised to match the input pair $(X_p, X_g)$ through a reconstruction of the combined latent representation:
\begin{equation}
\mathcal{L}_{\text{Diff}} =
\left\| \hat{X}_{\text{concat}} - X_{\text{concat}} \right\|_2^2,
\label{eq:x_concat}
\end{equation}
where $X_{\text{concat}} = \text{Concat}_{\text{height}}(X_p, X_g)$ serves as the target image. This supervision enforces global structure preservation and ensures that garment textures and positions are faithfully retained in the final output.

\vspace{0.5em}
\noindent\textbf{Feature Fusion in each layer.}  
In addition to latent concatenation, TryonNet integrates garment semantics from two complementary sources: (1) multi-scale garment features $\mathbf{F}_g^{(i)}$ extracted from GarmentNet’s self-attention layers, and (2) visual key--value pairs $(\mathbf{K}_d, \mathbf{V}_d)$ derived from the Light-Adapter. 
At each self-attention layer $i$ of TryonNet's U-Net backbone, the corresponding garment feature $\mathbf{F}_g^{(i)}$ is concatenated with the intermediate hidden state $\mathbf{H}_{\text{Tryon}}^{(i)}$ prior to the self-attention operation, forming an augmented representation $\mathbf{H}_{\text{aug}}^{(i)}$. This enables the model to jointly attend to both garment-specific structure and person-specific context across spatial scales, thereby improving alignment accuracy and enhancing the fidelity of garment details across layers.

In parallel, cross-attention is computed using two distinct branches of key--value pairs: one from the text prompt encoder $(\mathbf{K}_{\text{text}}, \mathbf{V}_{\text{text}})$ and one from the Light-Adapter $(\mathbf{K}_d, \mathbf{V}_d)$. Following the fusion strategy of IP-Adapter~\cite{ye_ip-adapter_2023}, the output attention features are computed as:
\begin{equation}
\mathbf{Z} = \text{Attention}(\mathbf{Q}, \mathbf{K}_{\text{text}}, \mathbf{V}_{\text{text}}) + \text{Attention}(\mathbf{Q}, \mathbf{K}_d, \mathbf{V}_d),
\end{equation}
where $\mathbf{Q}$ is the query projected from the hidden state. 
This dual-branch conditioning allows TryonNet to integrate textual guidance and visual garment semantics in a balanced manner, enhancing control over garment appearance and ensuring consistent garment-to-body alignment.

\begin{figure*}[ht]
    \centering
    \includegraphics[width=0.97\textwidth]{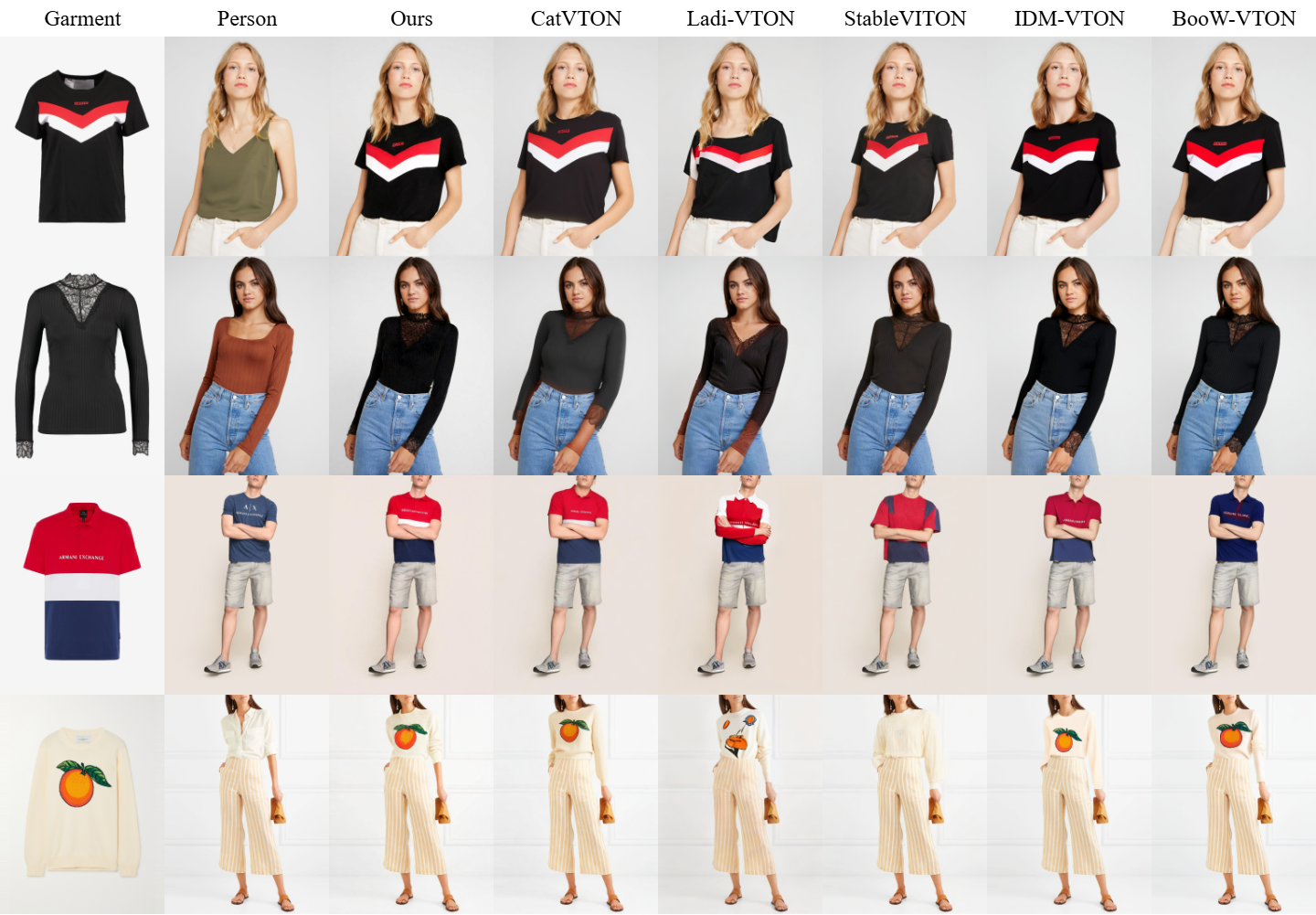}
        \caption{Qualitative comparison on VITON-HD and DressCode between our approach and server-side baselines. Zoom in for finer details.}
    \label{fig:Qualitative_comparison}
    \vspace{-1.em}
\end{figure*}

\subsection{Training Objective}
\label{sec:method:training}

All loss components described above are jointly optimized under a unified training objective tailored to the Feature-Guided Adversarial (FGA) framework. Each module is supervised by a set of objectives aligned with its functional role within the overall TGT architecture (Fig.~\ref{fig:Mobile-VTON}).

\vspace{0.5em}
\noindent\textbf{GarmentNet Losses.}  
GarmentNet is optimized using two objectives:
(i) a feature-level distillation loss $\mathcal{L}_{\text{feature}}^{\text{G}}$ that aligns its generative behavior with TeacherNet via score matching (Eq.~\ref{eq:dmg_final}), and
(ii) a trajectory consistency loss $\mathcal{L}_{\text{cons}}$ that ensures semantic stability of garment features across diffusion steps (Eq.~\ref{eq:cons}).
The total GarmentNet objective is:
\begin{equation}
\mathcal{L}_{\text{GarmentNet}} = \lambda_1 \mathcal{L}_{\text{feature}}^{\text{G}} + \lambda_2 \mathcal{L}_{\text{cons}}.
\end{equation}

\vspace{0.5em}
\noindent\textbf{TryonNet Losses.}  
TryonNet is optimized using three complementary supervision signals: (i) a feature-level distillation loss $\mathcal{L}_{\text{feature}}^{\text{T}}$ that aligns its generative score estimates with those of the teacher network (Eq.~\ref{eq:dmg_final}), (ii) an adversarial loss $\mathcal{L}_{\text{GAN}}$ that encourages photorealistic synthesis through a lightweight discriminator (Eq.~\ref{eq:gan}), and (iii) a garment-aware reconstruction loss $\mathcal{L}_{\text{Diff}}$ that enforces spatial alignment and preserves fine-grained garment details (Eq.~\ref{eq:x_concat}).
The final training objective for TryonNet is defined as:
\begin{equation}
\mathcal{L}_{\text{TryonNet}} = \mathcal{L}_{\text{Diff}} + \lambda_1 \mathcal{L}_{\text{feature}}^{\text{T}} + \lambda_3 \mathcal{L}_{\text{GAN}}.
\end{equation}

\vspace{0.5em}
\noindent\textbf{Final Objective.}  
The total loss for training \textsc{Mobile-VTON} integrates the objectives of both sub-networks:
\begin{equation}
\mathcal{L}_{\text{total}} = \mathcal{L}_{\text{GarmentNet}} + \mathcal{L}_{\text{TryonNet}},
\end{equation}
where $\lambda_1$, $\lambda_2$, and $\lambda_3$ are hyperparameters balancing distillation, consistency, and realism. This unified objective enables both GarmentNet and TryonNet to learn complementary, garment-aware representations for high-quality virtual try-on, without requiring large-scale pretraining.

\section{Experiments}

\begin{table*}[ht]
\centering

\resizebox{1\textwidth}{!}{
\begin{tabular}{lccccclccccclcc}
\hline
\textbf{Dataset} & \multicolumn{5}{c}{\textbf{VITON-HD}} &  & \multicolumn{5}{c}{\textbf{DressCode Upper-body}} &  & Memory & On \\ \cline{2-6} \cline{8-12} \cline{14-15} 
\textbf{Method} & LPIPS↓ & SSIM↑ & \multicolumn{1}{l}{CLIP-I↑} & FID↓ UN & KID↓ UN &  & LPIPS↓ & SSIM↑ & \multicolumn{1}{l}{CLIP-I↑} & FID↓ UN & KID↓ UN &  & Usage(G) & Mobile \\ \hline
\multicolumn{15}{c}{\cellcolor[HTML]{EFEFEF}\textbf{Server-side Mask-based methods}} \\ \hline
\textbf{SD-VITON \cite{shim_towards_2024}} & 0.104 & \cellcolor[HTML]{FFFC9E}\uline{0.890} & 0.831 & 9.857 & 1.450 &  & 0.201 & 0.858 & 0.555 & 50.755 & 38.331 &  & - & \ding{55} \\
\textbf{LaDI-VTON \cite{morelli_ladi-vton_2023}} & 0.166 & 0.873 & 0.819 & 9.386 & 1.590 &  & 0.157 & 0.905 & 0.773 & 22.689 & 2.580 &  & 7.63 & \ding{55} \\
\textbf{StableVITON \cite{kim_stableviton_2023}} & 0.142 & 0.875 & 0.838 & 9.371 & 1.990 &  & 0.113 & 0.910 & 0.844 & 19.712 & 2.149 &  & \color[HTML]{000000} 13.84 & \ding{55} \\
\textbf{IDM–VTON \cite{choi_improving_2024}} & \cellcolor[HTML]{FFFC9E}\uline{0.102} & 0.868 & \cellcolor[HTML]{FFCE93}\textbf{0.875} & \cellcolor[HTML]{FFFC9E}\uline{9.156} & \cellcolor[HTML]{FFFC9E}\uline{1.242} &  & 0.065 & 0.920 & \cellcolor[HTML]{FFCE93}\textbf{0.870} & \cellcolor[HTML]{FFFC9E}\uline{11.852} & \cellcolor[HTML]{FFFC9E}\uline{1.181} &  & 18.47 & \ding{55} \\ \hline
\multicolumn{15}{c}{\cellcolor[HTML]{EFEFEF}\textbf{Server-side Mask-Free methods}} \\ \hline
\textbf{CatVTON \cite{chong_catvton_2024}} & 0.161 & 0.872 & 0.832 & 11.275 & 2.847 &  & 0.0919 & 0.906 & 0.856 & 14.507 & 2.511 &  & \cellcolor[HTML]{FFFC9E}\uline{5.80} & \ding{55} \\
\textbf{BooW-VTON \cite{zhang_boow-vton_2024}} & 0.107 & 0.864 & \cellcolor[HTML]{FFFC9E}\uline{0.852} & \cellcolor[HTML]{FFCE93}\textbf{9.036} & \cellcolor[HTML]{FFCE93}\textbf{1.011} &  & \cellcolor[HTML]{FFCE93}\textbf{0.0513} & \cellcolor[HTML]{FFFC9E}\uline{0.928} & \cellcolor[HTML]{FFFC9E}\uline{0.869} & \cellcolor[HTML]{FFCE93}\textbf{11.361} & \cellcolor[HTML]{FFCE93}\textbf{1.043} &  & 18.47 & \ding{55} \\ \hline
\multicolumn{15}{c}{\cellcolor[HTML]{EFEFEF}\textbf{Mobile-side Mask-Free methods}} \\ \hline
\textbf{Ours} & \cellcolor[HTML]{FFCE93}\textbf{0.088} & \cellcolor[HTML]{FFCE93}\textbf{0.893} & 0.833 & 10.211 & 2.023 &  & \cellcolor[HTML]{FFFC9E}\uline{0.053} & \cellcolor[HTML]{FFCE93}\textbf{0.935} & 0.845 & 12.775 & 1.917 &  & \cellcolor[HTML]{FFCE93}\textbf{2.84} & \ding{51} \\ \hline
\end{tabular}
}
\caption{Quantitative comparison of models trained and evaluated on VITON-HD and DressCode upper-body datasets. \textbf{Bold} and \underline{underline} indicate the best and the second best results, respectively. ``UN'' denotes the unpaired setting. FID score is multiplied by 100.}
\label{tab:hd-result}
\vspace{-1.0em}
\end{table*}

\begin{table}[t]
\centering

\resizebox{1\linewidth}{!}{
\begin{tabular}{lccccc}
\hline
\textbf{Dataset} & \multicolumn{5}{c}{\textbf{VITON-HD In-the-wild}} \\ \hline
\textbf{Method} & LPIPS↓ & SSIM↑ & CLIP-I↑ & FID↓ UN & KID↓ UN \\ \hline
\multicolumn{6}{c}{\cellcolor[HTML]{EFEFEF}\textbf{Server-side Mask-based method}} \\ \hline
\textbf{SD-VITON} & 0.174 & 0.857 & 0.804 & 17.151 & 5.326 \\
\textbf{StableVITON} & 0.201 & 0.863 & 0.848 & 12.983 & 2.804 \\
\textbf{IDM–VTON} & \cellcolor[HTML]{FFFC9E}\uline{0.137} & 0.832 & \cellcolor[HTML]{FFCE93}\textbf{0.931} & \cellcolor[HTML]{FFFC9E}\uline{11.373} & \cellcolor[HTML]{FFFC9E}\uline{1.133} \\ \hline
\multicolumn{6}{c}{\cellcolor[HTML]{EFEFEF}\textbf{Server-side Mask-free method}} \\ \hline
\textbf{CatVTON} & 0.139 & \cellcolor[HTML]{FFCE93}\textbf{0.881} & \cellcolor[HTML]{FFFC9E}\uline{0.921} & 12.118 & 1.736 \\
\textbf{BooW-VTON} & 0.148 & 0.864 & 0.842 & \cellcolor[HTML]{FFCE93}\textbf{9.256} & \cellcolor[HTML]{FFCE93}\textbf{1.111} \\ \hline
\multicolumn{6}{c}{\cellcolor[HTML]{EFEFEF}\textbf{Mobile-side Mask-free method}} \\ \hline
\textbf{Ours} & \cellcolor[HTML]{FFCE93}\textbf{0.133} & \cellcolor[HTML]{FFFC9E}\uline{0.867} & 0.902 & 12.095 & 1.503 \\ \hline
\end{tabular}
}
\caption{Quantitative comparison of models on the VITON-HD In-the-Wild dataset. ``UN'' indicated the unpaired setting. FID score is multiplied by 100.}\label{tab:viton_wild}

\vspace{-1.em}
\end{table}

\subsection{Experiment Setup}

\noindent\textbf{Baselines.} We compare our mobile-side, mask-free method \textsc{Mobile-VTON} against both server-side, mask-based and mask-free state-of-the-art approaches. Mask-based baselines include SD-VITON~\cite{shim_towards_2024}, LaDI-VTON~\cite{morelli_ladi-vton_2023}, StableVITON~\cite{kim_stableviton_2023}, and IDM-VTON~\cite{choi_improving_2024}, which require segmentation masks during inference. Mask-free baselines such as CatVTON~\cite{chong_catvton_2024} and BooW-VTON~\cite{zhang_boow-vton_2024} rely entirely on diffusion-based generation without mask supervision. We follow a standard unpaired evaluation protocol and generate try-on images at \(1024 \times 768\) resolution, where supported. For lower-resolution outputs (e.g., \(512 \times 384\)), we apply bilinear interpolation or Real-ESRGAN~\cite{wang_real-esrgan_2021} to ensure consistent output size and quality across methods.

\vspace{0.5em}
\noindent\textbf{Evaluation Datasets.}  
We evaluate the performance of \textsc{Mobile-VTON} on three datasets: VITON-HD~\cite{choi_viton-hd_2021}, DressCode~\cite{morelli_dress_2022}, and an In-the-Wild variant of VITON-HD. The third dataset is constructed following the protocol introduced in BooW-VITON~\cite{zhang_boow-vton_2024}, enabling evaluation under real-world conditions. VITON-HD contains 13,679 image–garment pairs, consisting of frontal-view images of female subjects and their corresponding upper-body clothing. Following established practice~\cite{morelli_ladi-vton_2023, gou_taming_2023, kim_stableviton_2023, choi_improving_2024, wan_improving_2024}, we use 11,647 pairs for training and 2,032 for testing in both the original and In-the-Wild versions. DressCode includes 15,366 image–garment pairs focused on upper-body try-on. According to the official protocol, we select 1,800 samples from the upper-body subset for evaluation. All experiments are conducted at a resolution of \(1024 \times 768\) to ensure comparability with existing methods.

\vspace{0.5em}
\noindent\textbf{Evaluation Metrics.}  
We assess \textsc{Mobile-VTON} in both unpaired and paired evaluation settings, following standard protocols in previous virtual try-on studies~\cite{morelli_ladi-vton_2023, choi_viton-hd_2021, choi_improving_2024, wan_ted-viton_2024, wan_mf-viton_2025, zhang_boow-vton_2024, chong_catvton_2024}. In the unpaired setting, where the input garment differs from the original, we evaluate image realism and distributional similarity using Fréchet Inception Distance (FID)~\cite{heusel_gans_2018} and Kernel Inception Distance (KID)~\cite{kim_u-gat-it_2020}. This setup closely reflects real-world application scenarios. In the paired setting, where the target garment matches the original, we measure visual fidelity using three metrics: Structural Similarity Index (SSIM)~\cite{wang_image_2004} for structural consistency, Learned Perceptual Image Patch Similarity (LPIPS)~\cite{zhang_unreasonable_2018} for perceptual similarity, and CLIP image similarity (CLIP-I)~\cite{hessel_clipscore_2021} for semantic alignment.

\vspace{0.5em}
\noindent\textbf{Implementation Details}  
All experiments were conducted using 8 NVIDIA A100 GPUs (80GB). As the official implementation of SnapGen~\cite{chen_snapgen_2025} was not publicly available, we re-implemented the architecture based on the descriptions provided in the original paper. We adapted the model to the downstream \textsc{Mobile-VTON} task using a two-stage training process. In the first stage, the model was trained for 140 epochs on a merged dataset combining the upper-body subsets of DressCode~\cite{morelli_dress_2022} and VITON-HD~\cite{choi_viton-hd_2021}. In the second stage, it was fine-tuned for 100 additional epochs on the higher-quality DressCode subset to further improve consistency and quality. All training stages used a batch size of 256 and the AdamW optimizer~\cite{kingma_adam_2017} with a learning rate of $1\text{e}{-4}$ in the first stage and $5\text{e}{-5}$ in the second. The optimizer parameters were set to $\beta_1 = 0.9$ and $\beta_2 = 0.999$. The overall loss function was weighted using $\lambda_1 = 1\text{e}{-2}$, $\lambda_2 = 0.5$, and $\lambda_3 = 5\text{e}{-3}$ throughout both stages.

\subsection{Qualitative Results}

Figure~\ref{fig:cover_image} compares visual results on the VITON-HD In-the-Wild benchmark. Compared to server-based baselines such as CatVTON, LaDI-VTON, StableVITON, IDM-VTON, and BooW-VTON, our method produces competitive texture details and stable garment boundaries, with improved alignment in challenging regions such as logos and sleeves. Figure~\ref{fig:Qualitative_comparison} shows results across varied poses and garment types on DressCode and VITON-HD. \textsc{Mobile-VTON} consistently preserves garment structure and fabric appearance under pose and style variations. It accurately captures details such as lace, color transitions, and fabric deformation, where other methods often introduce smoothing or misalignment. Despite its compact size (0.41B parameters), \textsc{Mobile-VTON} delivers high visual fidelity while being the only method that runs fully on mobile devices. Competing methods typically contain 2--7 times more parameters and require server-class GPUs for inference. These comparisons highlight both the visual quality and practical deployability of our approach for real-world try-on applications.

\subsection{Quantitative Results}

As shown in Table~\ref{tab:hd-result}, our method achieves the best overall SSIM and the second-lowest LPIPS on the DressCode Upper-body dataset, indicating strong structural preservation and perceptual quality. On VITON-HD, \textsc{Mobile-VTON} ranks among the top performers across LPIPS, SSIM, and CLIP-I, while maintaining competitive FID and KID scores. Table \ref{tab:viton_wild} further reports results on the VITON-HD In-the-Wild benchmark. Our method performs on par with or better than server-based models in LPIPS and SSIM, while using significantly less memory. As reported in Table~\ref{tab:hd-result}, \textsc{Mobile-VTON} requires only 2.84 GB of memory, compared to 5 to 18 GB baselines. Importantly, most server-side baselines operate using segmentation masks to restrict generation to the garment region. In contrast, our model is completely mask free and must synthesize the entire image, including the body, clothing, and background, without any spatial priors. This makes realism-oriented metrics such as FID and KID considerably more difficult. Despite this, \textsc{Mobile-VTON} still achieves competitive performance, demonstrating its robustness, generalization ability, and practical deployability in real-world settings.

\begin{figure}[t]
    \centering
    \includegraphics[width=1.0\linewidth]{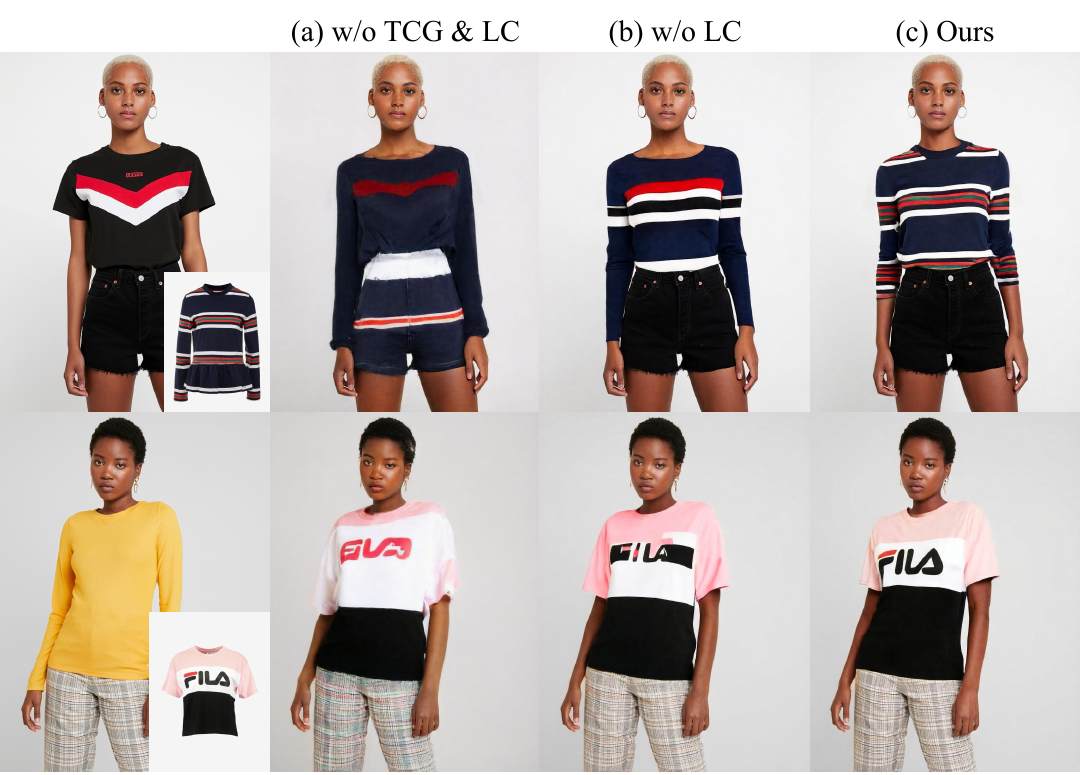}
    \caption{Comparison of virtual try-on results with and without the TCG and LC. Zoom in for more details.}
    \label{fig:ablation}
\vspace{-1.em}
\end{figure}

\begin{table}[t]
    \centering
    
    \resizebox{1\linewidth}{!}{
        \begin{tabular}{cc|ccccc}
        \hline
        \textbf{TCG} & \textbf{LC} & \textbf{LPIPS↓} & \textbf{SSIM↑} & \textbf{CLIP-I↑} & \textbf{FID↓ UN} & \textbf{KID↓ UN} \\ \hline
        \ding{55} & \ding{55} & 0.119 & 0.874 & 0.798 & 11.231 & 3.136 \\
        \ding{51} & \ding{55} & 0.111 & 0.879 & 0.805 & 10.814 & 2.744 \\
        \ding{51} & \ding{51} & \textbf{0.088} & \textbf{0.893} & \textbf{0.833} & \textbf{10.211} & \textbf{2.023} \\ \hline
        \end{tabular}
    }
    \caption{Quantitative comparison of models trained on the VITON-HD dataset with components. ``UN'' indicated the unpaired setting. FID score is multiplied by 100.}
    \label{tab:result_itw}
    \vspace{-1.5em}
\end{table}

\subsection{Ablation Study}

\noindent\textbf{Effect of Trajectory-Consistent GarmentNet (TCG).}  
As shown in Table~\ref{tab:result_itw}, adding the TCG module improves all metrics on the VITON-HD dataset. LPIPS drops from 0.119 to 0.111, SSIM increases from 0.874 to 0.879, and CLIP-I improves from 0.798 to 0.805. FID and KID also decrease, indicating better image realism and distributional alignment. Figure~\ref{fig:ablation} shows a visual comparison between models trained with and without TCG. Without TCG (column a), the model struggles to preserve spatial correspondence between garment colors and body structure. Critical elements such as stripes and logos appear faded, misaligned, or missing. For example, the red chevron in the top row is broken and misplaced, while the ``FILA" logo in the second row is distorted and lacks contrast. In contrast, enabling TCG (column b) improves garment-to-body alignment. The model better preserves original garment colors and their spatial layout, producing outputs that more closely reflect the source design. Logos and stripe patterns are sharper and better positioned, indicating that TCG improves semantic stability and helps the network maintain color-to-location consistency across poses.

\vspace{0.5em}
\noindent\textbf{Effect of Latent Concatenation (LC).}
Building on TCG, adding LC further improves all metrics: LPIPS drops to 0.088, SSIM increases to 0.893, and CLIP-I rises to 0.833. FID and KID also decrease, indicating higher overall realism. This performance gain stems from LC’s ability to directly inject garment information into TryonNet, enabling it to synthesize garments from scratch rather than relying on pretraining. As shown in Figure~\ref{fig:ablation} (column c vs. b), LC enhances garment generation quality and structural alignment. In the top row, stripe placement, sleeve length, and multi-color patterns better match the reference garment. In the second row, the pink-black-white color blocking and the ``FILA" logo are rendered with improved clarity and placement. These results show that LC complements TCG by providing TryonNet with explicit garment geometry and appearance cues, overcoming the lack of large-scale pretraining and achieving final, high-fidelity synthesis.

\section{Conclusion}
\label{sec:conclusion}

We introduced \textsc{Mobile-VTON}, a unified distillation-driven framework that enables high-fidelity virtual try-on directly on mobile devices. 
Built on the proposed TGT architecture, our approach transfers both generative behaviors and perceptual priors from a large diffusion teacher to lightweight student networks. 
GarmentNet extracts temporally consistent garment features, while TryonNet aligns garments with person representations through structured attention and latent concatenation. 
A lightweight Light-Adapter further facilitates efficient garment conditioning, and the training objective integrates distillation, adversarial supervision, and trajectory consistency. 
Overall, \textsc{Mobile-VTON} demonstrates that diffusion-based try-on systems can achieve strong visual fidelity while remaining efficient enough for fully on-device deployment.

{
    \small
    \bibliographystyle{ieeenat_fullname}
    \bibliography{references}
}

\clearpage
\setcounter{page}{1}
\maketitlesupplementary


\section*{A. On-Device Deployment Details}

We evaluate the practical feasibility of \textsc{Mobile-VTON} under a fully on-device setting. The model is deployed on a Xiaomi 17 Pro Max equipped with a Snapdragon 8 Gen 5 chipset and Qualcomm Hexagon NPU. All experiments are conducted in Airplane Mode to ensure that no external computation or network communication is involved.

The full inference pipeline runs entirely on the mobile device. Model weights are preloaded onto the NPU, incurring negligible runtime loading overhead. GarmentNet and TryOnNet execute in parallel on the NPU, while the VAE operates on the CPU during encoding and is transferred to the NPU only for the final decoding stage.
The end-to-end inference time for generating one full-resolution ($1024\times768$) try-on result is approximately 80 seconds on the mobile NPU. This latency corresponds to a complete diffusion pipeline executed without step reduction, pruning, or system-level acceleration techniques.
We adopt INT8 quantization during inference, as most Android NPUs (including Qualcomm Hexagon) require INT8 execution. In contrast, Apple A-series NPUs support BF16 inference (as reported in SnapGen), indicating that the proposed architecture is compatible with both major mobile hardware ecosystems. These results demonstrate the feasibility of deploying diffusion-based virtual try-on models fully on-device without reliance on cloud-based computation.

\section*{B. Additional Ablation Study}

To better understand the contribution of each component in \textsc{Mobile-VTON}, we conduct controlled ablation experiments on the DressCode upper-body setting. Specifically, we evaluate the impact of (a) the Latent Alignment (LA) module, (b) Trajectory-Consistent Garment supervision (TCG), and (c) the distillation training strategy.

As shown in Table~\ref{tab:ablation_study}, removing the LA module significantly degrades perceptual metrics (LPIPS increases from 0.088 to 0.168, SSIM drops from 0.893 to 0.827), indicating that latent-level garment alignment is crucial for preserving fine-grained appearance details. 
Removing TCG supervision also leads to consistent degradation across all metrics, suggesting that enforcing trajectory-consistent feature transfer across diffusion timesteps helps stabilize garment representation and reduces structural drift.
Most notably, training without distillation results in severe performance collapse (FID increases from 10.211 to 113.59), demonstrating that direct training from scratch without teacher guidance fails to converge to a meaningful solution under the lightweight model capacity.

\begin{figure}[t]
    \centering
    \includegraphics[width=\linewidth]{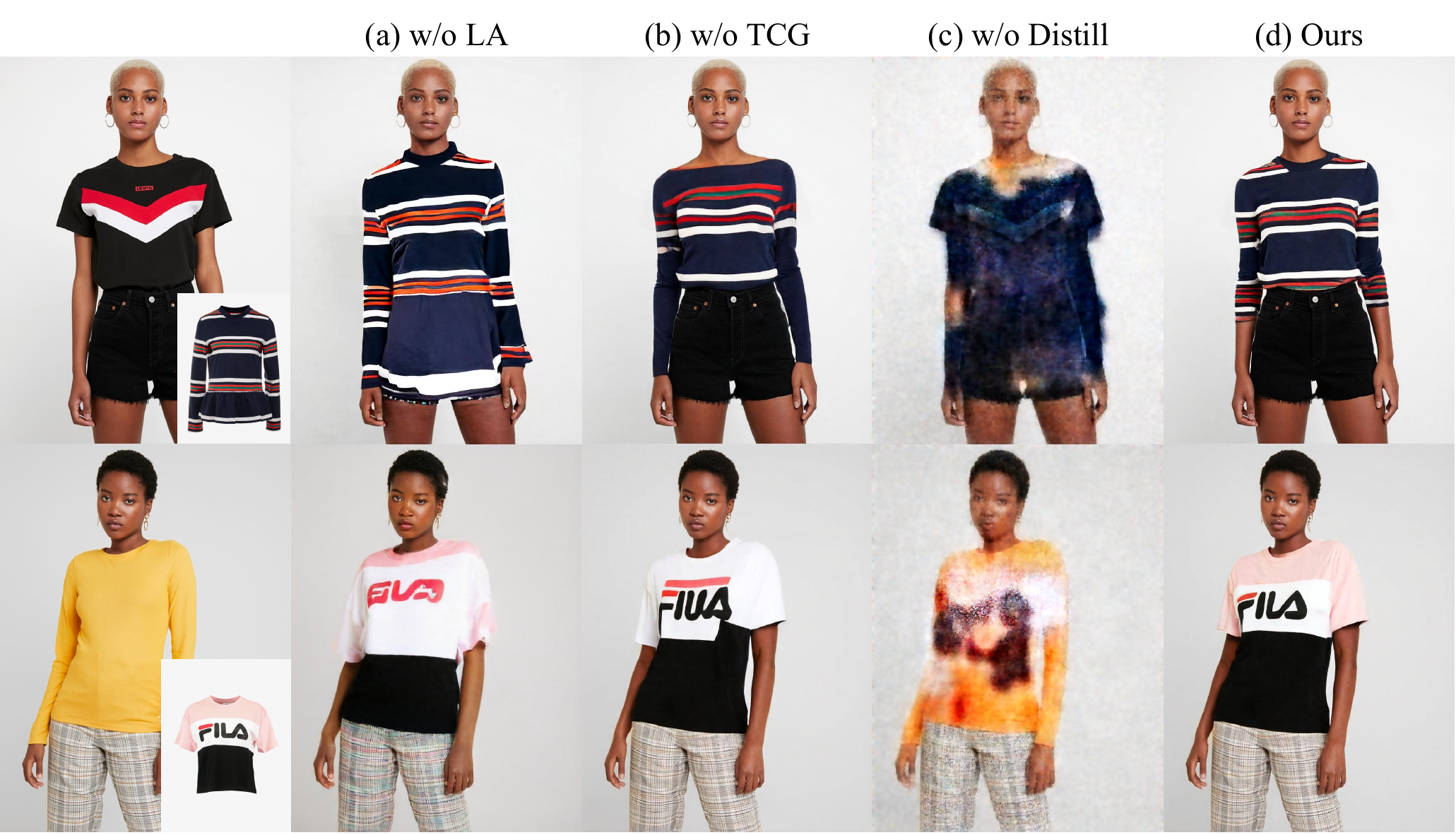}
    \caption{Qualitative ablation results on the DressCode upper-body test set. Removing key components degrades garment detail, structural consistency, or overall synthesis quality.}
    \label{fig:ablation_study}
    \vspace{-1em}
\end{figure}

Qualitative comparisons in Fig.~\ref{fig:ablation_study} further corroborate these observations. LA improves garment texture fidelity, TCG enhances structural coherence, and distillation provides essential supervision for stable training. Together, these components form a complementary design in which each contributes to model stability and perceptual quality.

\begin{table}[t]
\centering
\caption{Quantitative ablation results on the DressCode upper-body test set (UN setting). Each component contributes to improved perceptual quality and distribution alignment.}
\resizebox{1\linewidth}{!}{
\begin{tabular}{lccccc}
\hline
\textbf{Method} & \textbf{LPIPS↓} & \textbf{SSIM↑} & \textbf{CLIP-I↑} & \textbf{FID↓ (UN)} & \textbf{KID↓ (UN)} \\ \hline
Ours & \textbf{0.088} & \textbf{0.893} & \textbf{0.833} & \textbf{10.211} & \textbf{2.023} \\
\;\;\; w/o TCG & 0.108 & 0.881 & 0.804 & 10.914 & 2.939 \\
\;\;\; w/o LA & 0.168 & 0.827 & 0.704 & 13.652 & 5.091 \\
\;\;\; w/o Distill & 0.354 & 0.788 & 0.438 & 113.59 & 106.66 \\ \hline
\end{tabular}
}
\label{tab:ablation_study}
\vspace{-1em}
\end{table}



\section*{C. Discussion on GarmentNet}

GarmentNet is trained under Trajectory-Consistent Garment (TCG) supervision, where real garment images are directly used as reconstruction targets across diffusion timesteps. 
This design provides explicit supervision from the true data distribution and promotes semantic consistency throughout the denoising trajectory.
Under this training setting, introducing an additional adversarial loss for GarmentNet did not yield measurable improvements in either quantitative or perceptual metrics, while increasing training complexity and optimization instability. 
We therefore do not incorporate a GAN loss into the final design.
This observation is in line with prior studies emphasizing the importance of training stability and supervision quality in preventing shortcut dependency and over-memorization in constrained models~\cite{lin2023eliminating,linover,linlayer}. 
Related work also highlights that carefully conditioned learning signals are essential for robustness and cross-modal generalization in generative systems~\cite{linunderstanding,lin2025force,zheng2026vii,xiang2026safety,hong2025adlift,huang2024machine,huang2025towards}.

\section*{D. Impact of Different Dataset Quality}

While prior studies consistently report better results from fine-tuning on the VITON-HD dataset, we observe the opposite trend with our lightweight \textsc{Mobile-VTON}: it achieves superior performance when fine-tuned on the DressCode dataset.
To investigate how dataset quality influences the performance of lightweight virtual try-on models, we conduct fine-tuning experiments on both VITON-HD and DressCode. All models are initially fine-tuned on a combined dataset consisting of both VITON-HD and DressCode, and are subsequently fine-tuned individually on each dataset. We perform evaluations on both test sets to assess generalization ability and cross-domain robustness.

As shown in Table~\ref{tab:dataset_comp}, our model achieves consistent improvements across all metrics when fine-tuned on DressCode, even when evaluated on the VITON-HD test set. For example, on VITON-HD, SSIM improves from 0.932 to 0.935, LPIPS drops by 9.3\%, and CLIP-I increases by 1.56\%, while realism-oriented metrics such as FID and KID also show notable gains. The performance gap is even larger when evaluating on the DressCode test set: SSIM improves by 1.82\%, LPIPS by 13.9\%, and KID by 9.5\%.

\begin{figure}[t]
    \centering
    \includegraphics[width=1.0\linewidth]{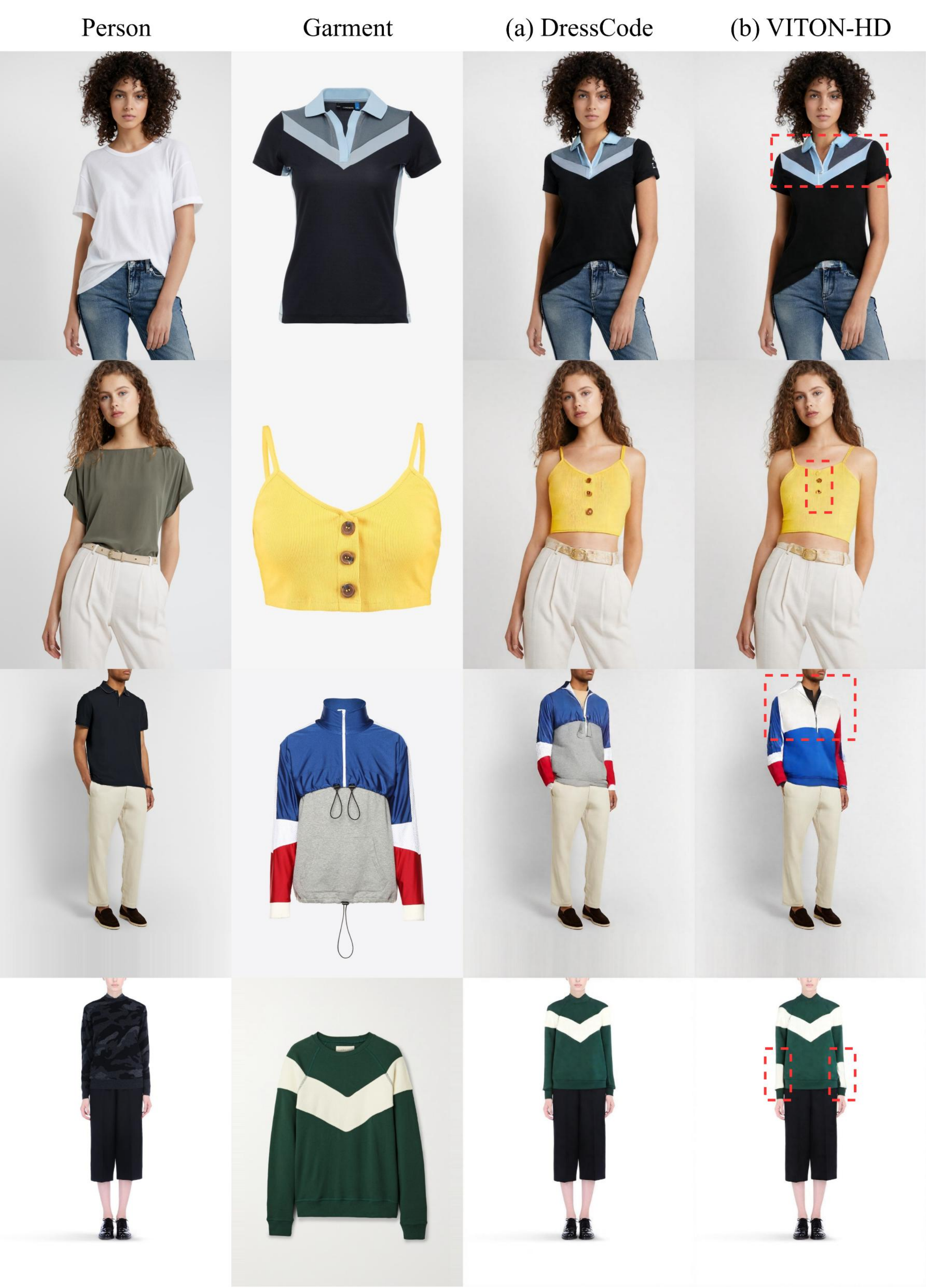}
    \caption{Qualitative comparison of try-on results fine-tuned on DressCode (a) and VITON-HD (b). DressCode-trained models produce sharper textures and more accurate alignment, especially in fine-grained regions (highlighted in red). Zoom in for details.}
    \label{fig:ablation_sup}
\end{figure}

Visual comparisons in Fig.~\ref{fig:ablation_sup} illustrate that models trained on DressCode yield sharper garment textures and more coherent alignment, particularly in fine-grained regions such as logos and sleeves. We attribute these consistent gains to the higher sensitivity of lightweight models to data quality. The DressCode dataset, with its uniform resolution, consistent garment framing, and clearer visual features, offers more stable learning signals, which are essential for on-device models with limited parameter budgets. In contrast, VITON-HD contains a mix of low- and high-quality images, with varying degrees of compression and pose ambiguity, which can degrade the training dynamics for small models. These findings underscore the importance of dataset curation and quality when designing practical, efficient mobile-based VTON systems.

\section*{E. Limitations}

A key limitation of \textsc{Mobile-VTON} lies in its difficulty in accurately reproducing garments with textual elements, such as logos, printed slogans, or brand names. These failures are often manifested as blurred, distorted, or partially missing characters in the generated try-on results. This issue primarily stems from two factors: first, the model is trained entirely from scratch without leveraging any large-scale pretraining on text-aware image corpora; second, garments containing prominent textual content are relatively rare in current virtual try-on datasets, which limits the model’s exposure to such patterns during training. As a result, the model lacks the capacity to establish robust visual-text associations and generalize to unseen text styles or layouts. This limitation is particularly evident in cases involving stylized fonts, curved text, or small printed labels. 

\begin{table}[t]
\centering
\caption{Comparison of \textsc{Mobile-VTON} fine-tuned on VITON-HD and DressCode upper-body. ``UN'' indicates unpaired setting; FID is multiplied by 100.}
\resizebox{1\linewidth}{!}{
\begin{tabular}{lccccc}
\hline
\textbf{Dataset} & \multicolumn{1}{l}{\textbf{LPIPS↓}} & \multicolumn{1}{l}{\textbf{SSIM↑}} & \multicolumn{1}{l}{\textbf{CLIP-I↑}} & \multicolumn{1}{l}{\textbf{FID↓ UN}} & \multicolumn{1}{l}{\textbf{KID↓ UN}} \\ \hline
\multicolumn{6}{c}{\cellcolor[HTML]{EFEFEF}\textbf{VITON-HD}} \\ \hline
\textbf{VITON-HD} & 0.102 & 0.877 & 0.818 & 10.558 & 2.685 \\
\textbf{DressCode} & \textbf{0.088} & \textbf{0.893} & \textbf{0.833} & \textbf{10.211} & \textbf{2.023} \\ \hline
\multicolumn{6}{c}{\cellcolor[HTML]{EFEFEF}\textbf{DressCode Upper-body}} \\ \hline
\textbf{VITON-HD} & 0.058 & 0.932 & 0.832 & 13.527 & 2.473 \\
\textbf{DressCode} & \textbf{0.053} & \textbf{0.935} & \textbf{0.845} & \textbf{12.775} & \textbf{1.917} \\ \hline
\end{tabular}
}
\label{tab:dataset_comp}
\vspace{-1.5em}
\end{table}

\begin{figure*}[t]
    \centering
    \includegraphics[width=0.94\textwidth]{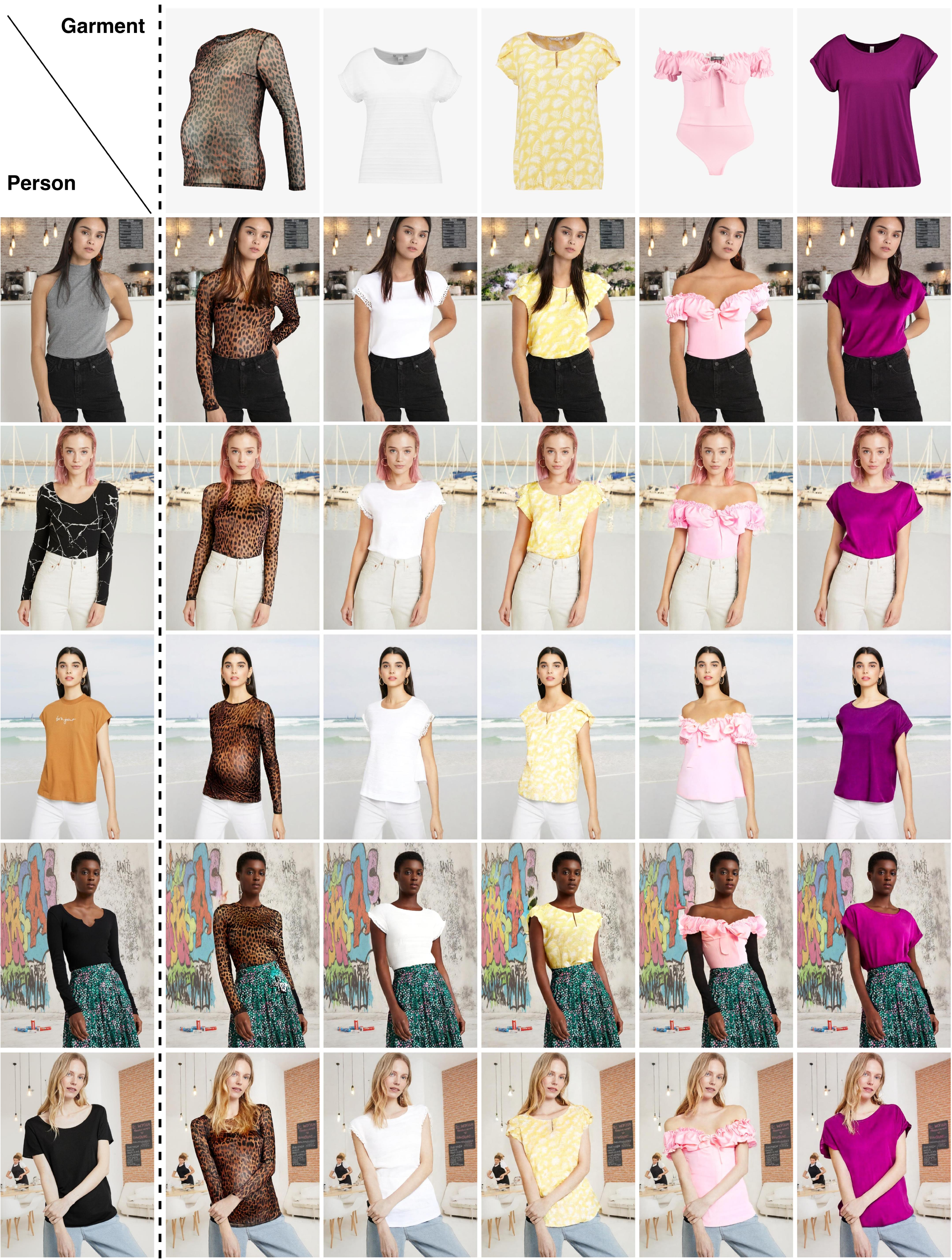}
        \caption{Try-on results on the VITON-HD In-the-Wild test set, produced by \textsc{Mobile-VTON}. Zoom in for finer details.}
    \label{fig:VITON}
\end{figure*}

\begin{figure*}[t]
    \centering
    \includegraphics[width=0.94\textwidth]{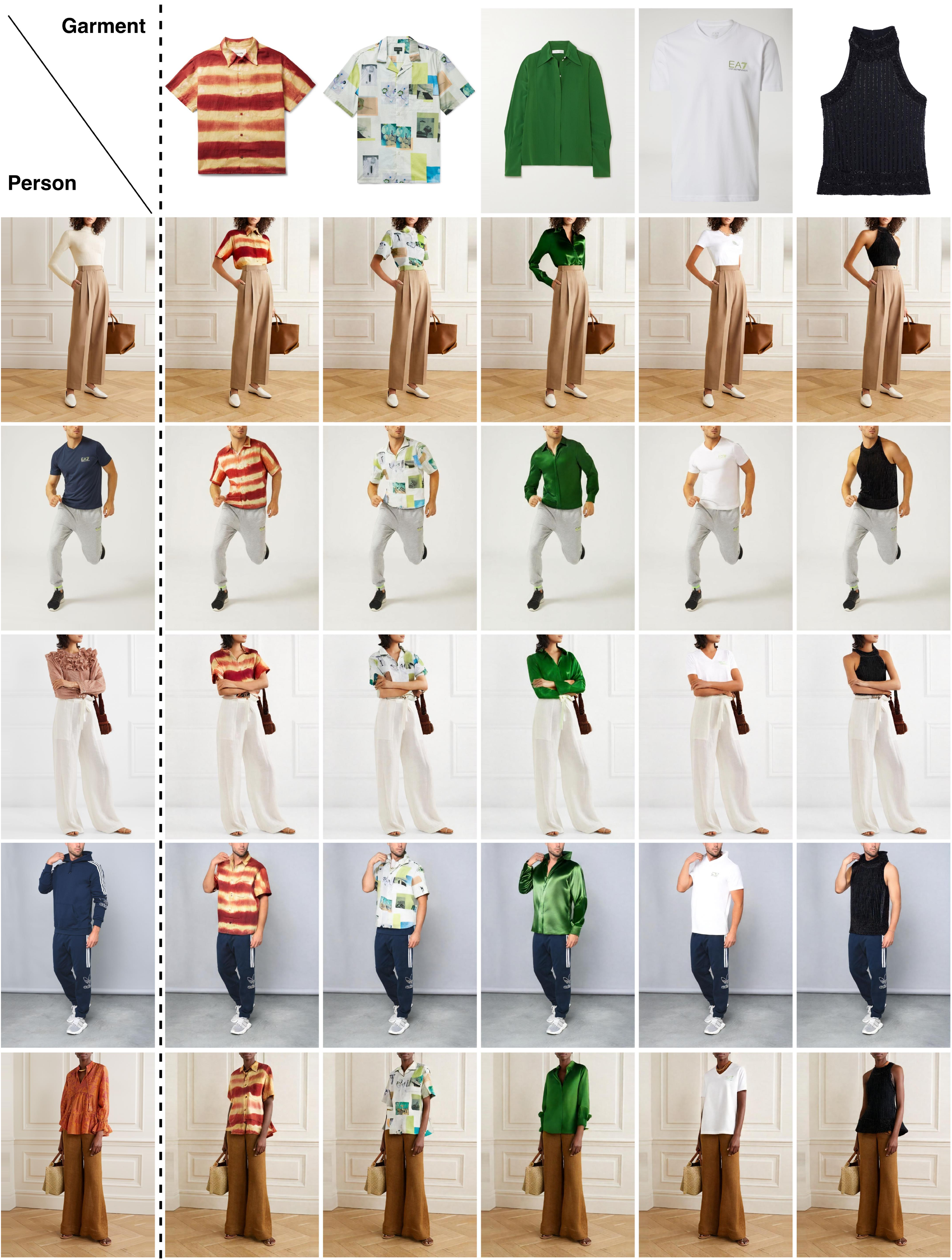}
        \caption{Try-on results on the DressCode test set, produced by \textsc{Mobile-VTON}. Zoom in for finer details.}
    \label{fig:VITON}
\end{figure*}

\begin{figure*}[t]
    \centering
    \includegraphics[width=0.94\textwidth]{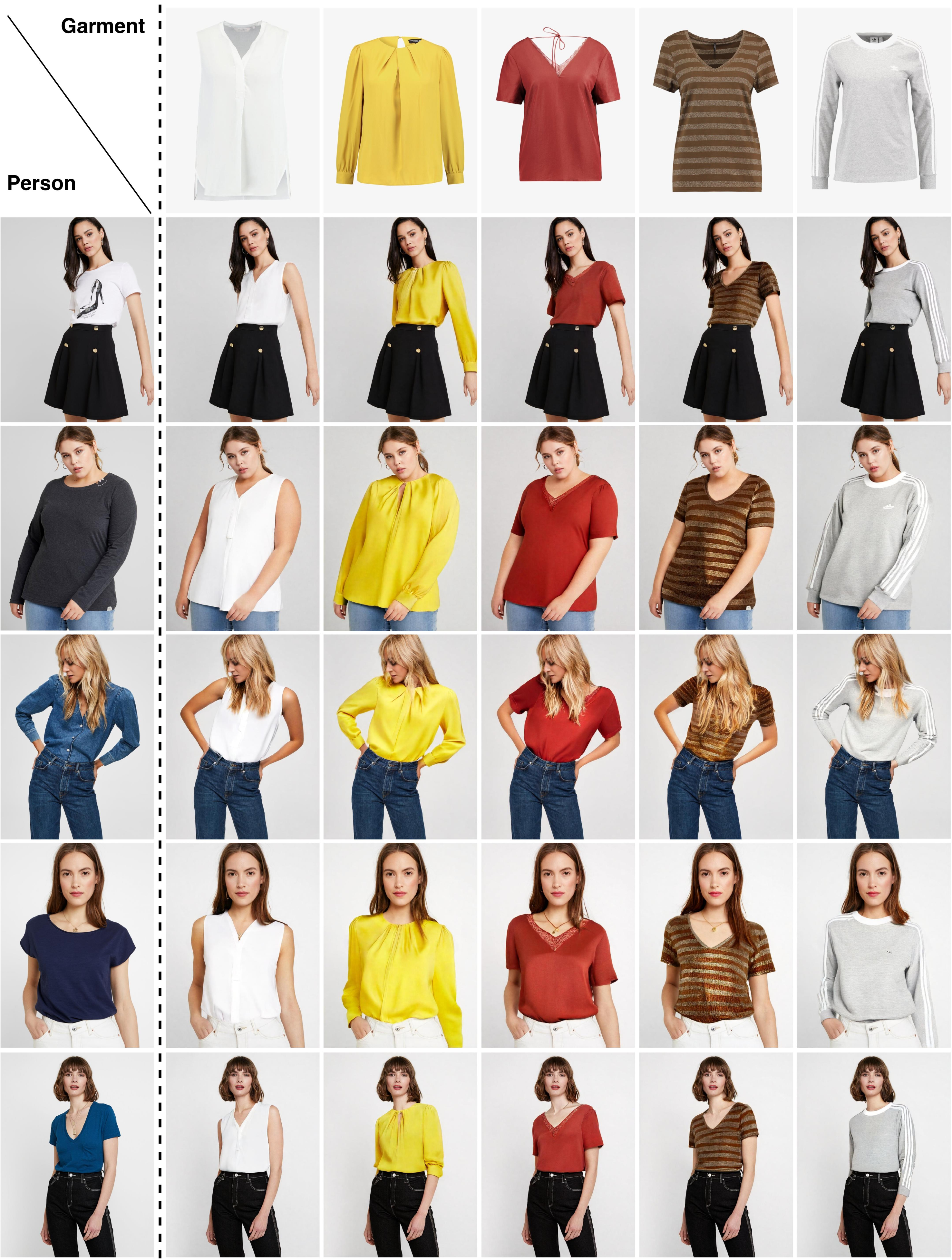}
        \caption{Try-on results on the VITON-HD test set, produced by \textsc{Mobile-VTON}. Zoom in for finer details.}
    \label{fig:VITON}
\end{figure*}

\end{document}